\documentclass[11pt,x11names]{article}
\usepackage{setspace}
\usepackage[]{misc/emnlp2023}
\def\anonymized#1{#1}

\usepackage{times}
\usepackage{xcolor}
\usepackage{graphicx}
\usepackage[T1]{fontenc}
\usepackage[utf8]{inputenc}
\usepackage{microtype}
\usepackage{booktabs}
\usepackage{todonotes}
\usepackage{amsmath,amssymb}
\usepackage{cleveref}
\usepackage{xspace}
\usepackage{soul} %
\usepackage{caption}
\usepackage{multirow,multicol}
\usepackage{ulem}
\usepackage{float}
\usepackage{array}
\usepackage{bm}
\usepackage{xfrac}
\usepackage{tcolorbox} %
\usepackage[shortlabels]{enumitem}
\usepackage[outline]{contour}%
\usepackage{tikz}
\usepackage{fontawesome}

\setlist[itemize]{noitemsep,left=0mm}

\usepackage{tabularx}

\crefname{lstlisting}{listing}{listings}
\Crefname{lstlisting}{Listing}{Listings}
\crefname{equ}{equation}{equations}
\Crefname{equ}{Equation}{Equations}
\Crefname{algorithm}{Algorithm}{Algorithms}
\crefname{example}{example}{examples}
\Crefname{example}{Example}{Examples}
\crefname{prompt}{prompt}{prompts}
\Crefname{prompt}{Prompt}{Prompts}
\DeclareCaptionType{example}[Example][List of examples]
\DeclareCaptionType{prompt}[Prompt][List of prompts]
\DeclareCaptionType{equ}[Equation][List of equations]

\usepackage{marginnote}

\definecolor{TodoColor}{rgb}{1,0.7,0.6}
\definecolor{TodoColor2}{rgb}{0.7,0.7,0.9}
\definecolor{TodoColor3}{rgb}{0.5,0.8,0.5}

\newcommand{\ACref}[1]{App.\hspace{0.1em}\Cref{#1}}

\newcommand{\mqmwmt}{MQM\textsuperscript{WMT}\xspace}
\newcommand{\esaspans}{ESA\textsubscript{spans}\xspace}
\newcommand{\dasqmwmt}{DA+SQM\textsuperscript{WMT}\xspace}

\newcommand{\hlc}[2][yellow]{{%
    \colorlet{foo}{#1}%
    \sethlcolor{foo}\hl{#2}}%
}

\tikzset{every node/.style={anchor=base, inner sep=2pt, outer sep=0}}

\makeatletter\def\Hy@Warning#1{}\makeatother
\let\svthefootnote\thefootnote
\newcommand\blankfootnote[1]{%
  \let\thefootnote\relax\footnotetext{#1}%
  \let\thefootnote\svthefootnote%
}

\graphicspath{{img/}}

\contourlength{0.2pt}
\title{
    \scalebox{1.1}{E}rror \scalebox{1.1}{S}pan \scalebox{1.1}{A}nnotation:\\
    A Balanced Approach for Human Evaluation of Machine Translation
}

\newcommand{\tstar}{$^{\bigstar}$}

\author{
    Tom Kocmi\tstar$^1$ \qquad
    Vilém Zouhar\tstar$^2$ \qquad    
    Eleftherios Avramidis$^3$  \qquad
    Roman Grundkiewicz$^1$ \\
    \bf
    Marzena Karpinska$^4$ \qquad
    Maja Popović$^5$ \qquad
    Mrinmaya Sachan$^2$ \qquad
    Mariya Shmatova$^6$
    \\[0.5em]
    \begin{tabular}{ccc}
    $^1$Microsoft &
    $^2$ETH Zurich &
    $^3$DFKI \\
    $^4$UMass Amherst&
    $^5$DCU \& IU  &
    $^6$Dubformer \\ &
     \\
    \end{tabular}\\[0.5em]
    {\tt\fontsize{9}{8}\selectfont 
     \href{mailto:tomkocmi@microsoft.com}{\color{black} tomkocmi@microsoft.com}
     \quad
     \href{mailto:vzouhar@inf.ethz.ch}{\color{black} vzouhar@inf.ethz.ch}
    }
}

\begin{document}

\maketitle

\maketitle

\begin{abstract}

High-quality Machine Translation (MT) evaluation relies heavily on human judgments.
Comprehensive error classification methods, such as Multidimensional Quality Metrics (MQM), are expensive as they are time-consuming and can only be done by experts, whose availability may be limited especially for low-resource languages.
On the other hand, just assigning overall scores, like Direct Assessment (DA), is simpler and faster and can be done by translators of any level, but is less reliable.
In this paper, we introduce Error Span Annotation (ESA), a human evaluation protocol which combines the continuous rating of DA with the high-level error severity span marking of MQM.
We validate ESA by comparing it to MQM and DA for 12 MT systems and one human reference translation (English to German) from WMT23. 
The results show that ESA offers faster and cheaper annotations than MQM at the same quality level, without the requirement of expensive MQM experts.
\end{abstract}

\blankfootnote{\anonymized{\tstar Equal contributions. Others alphabetically.}}
\blankfootnote{
\anonymized{\,$^0$Code \& collected data:\\
\null\hfill  \href{https://github.com/wmt-conference/ErrorSpanAnnotation}{\faGithub \ github.com/wmt-conference/ErrorSpanAnnotation} 
}}

\section{Introduction}\label{sec:introduction}

While automatic evaluation metrics are important and invaluable tools for rapid development of Machine Translation (MT) systems, human assessment remains the gold standard of translation quality \citep{kocmi-etal-2023-findings, freitag-etal-2023-results}.
The translation quality is conceptually measured through adequacy (preservation of the original meaning) and fluency (grammaticality of the translated text; \citealp{koehn-monz-2006-manual}), and sometimes through comprehension (how readable or understandable the translation is; \citealp{white-etal-1994-arpa}).

Annotators are usually asked to assign a score on a particular quality aspect.
Likert and 0--100 scale are often used for discrete and continuous scales.
The most popular scoring method in machine translation field in recent years is Direct Assessment \citep[DA;][]{graham-etal-2013-continuous}, which is used to portray a human assessment of MT quality in the \href{https://www2.statmt.org/wmt23/translation-task.html}{WMT shared tasks} since 2016.
Since 2022, the DA+SQM metric is used, namely direct assessment enriched with more objective Scalar Quality Metrics (SQM) guidelines \citep{kocmi-etal-2022-findings}. 

\begin{figure}[t]
\centering
\includegraphics[width=\linewidth]{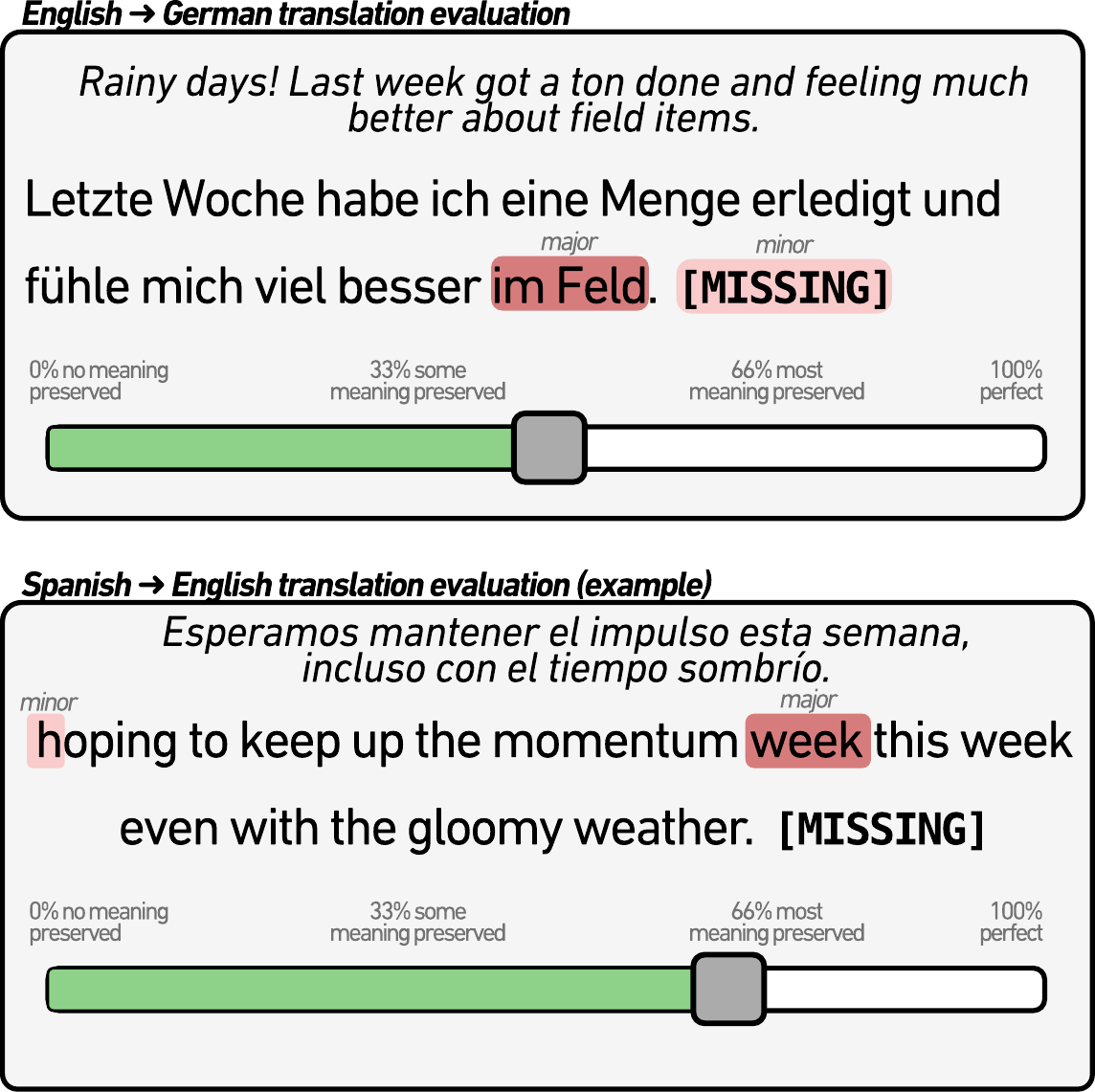}
\vspace{-7mm}
\caption{Stylized annotation user interface with Error Span Annotation (ESA). The annotator first marks errors with \hlc[red!20]{minor} and \hlc[red!50]{major} severity and then assigns a final score. This is more robust than asking for score directly.\footnotemark}
\label{fig:appraise_screenshot}
\vspace{-4mm}
\end{figure}
\footnotetext{Our experiments are on English$\rightarrow$German. In \Cref{fig:appraise_screenshot}, Spanish$\rightarrow$English is only an illustration for English-speaking readers. The first example, based on our data, omits \textit{Rainy days!} and incorrectly translates \textit{about field items} as \textit{im Feld} (\textit{=in the field}). The second example, for illustrative purposes, does not capitalize \textit{H} and mistakenly adds extra \textit{week}.}

Translation scores indicate the overall quality of a translation, but they can be subjective and do not provide details about the translation errors.
The usual way to overcome this drawback is error classification: asking the evaluators to mark each translation error and assign an error tag from a set of predefined categories, such as \textit{terminology} or \textit{style}.
In recent years, the dominant error classification protocol is the Multidimensional Quality Metrics \citep[MQM;][]{lommel-etal-2014-using, freitag-etal-2021-experts}.
MQM error classification is the standard human metric in the WMT Metrics shared task since 2021 \cite{freitag-etal-2021-results}.
While error classification provides interesting insights into the distribution of different types of errors, it requires much more time and effort, both for annotators and task organizers, who need to prepare the error taxonomy, annotation guidelines and training examples.

We present a new evaluation protocol based on highlighting errors and followed by assigning scores, Error Span Annotation (ESA), and compare it to the MQM error classification and DA+SQM scores. 
We compare MQM, DA+SQM and ESA annotations in parallel on a subset of English$\rightarrow$German machine translation outputs from WMT23.
We find that the proposed ESA protocol is faster and cheaper than MQM whilst providing the same usefulness in ranking MT systems.

\section{Related work}\label{sec:related}

Assigning overall scores was the very first method of manual MT evaluation \citep{alpac66,white-etal-1994-arpa}, where the evaluators assessed some or all of the translation quality criteria at once: adequacy, comprehensibility, and fluency.
\href{http://statmt.org/wmt06/}{The first WMT} (Workshop/Conference on Machine Translation) shared task in 2006 and the subsequent task in 2007 adopted this technique and used adequacy and fluency scores as official metrics \citep{koehn-monz-2006-manual,callison-burch-etal-2007-meta}.
Later, \citet{vilar-etal-2007-human} proposed binary ranking of two or more MT outputs, which became the official metric at WMT 2008 \cite{callison-burch-etal-2008-meta}.
It required less effort and showed better inter-annotator agreement than adequacy and fluency scores.
This remained the official WMT metric until 2017 when it was replaced by continuous Direct Assessment (DA).
DA \cite{graham-etal-2013-continuous} does not use discrete scales, but a continuous one between 0 and 100.
\citet{bojar-etal-2016-findings} scrutinized the quality criteria and recommended to focus on adequacy and use fluency to break ties only.
DA replaced ranking methods in 2017 \cite{bojar-etal-2017-findings} and since 2022 \citep{kocmi-etal-2022-findings} it is used with SQM guidelines \citep{freitag-etal-2021-experts} in a slightly modified version with more descriptive scale labels, which increased the inter-annotator agreement.

None of the described methods provides information about the erroneous or problematic parts of the translation.
An early work of \citet{vilar-etal-2006-error} analyzes errors in translation outputs assigning them to error classes from a predefined error typology. Most popular error typology recently is \href{https://themqm.org/error-types-2/typology/}{Multidimensional Quality Metrics} \citep[MQM;][]{lommel-etal-2014-using,hreval2018,freitag-etal-2021-experts}, which is used in WMT metrics task since 2021 \cite{freitag-etal-2021-results}.

Several error span marking methods have been proposed recently \cite{kreutzer-etal-2020-correct,popovic-2020-informative} as a less demanding error annotation approach than error classification.
While it does not provide the fine-grained details about different error classes, it still gives the information about the position and amount of errors, and also enables further fine-grained analysis on the annotated data, if necessary (e.g. classification of already marked errors, identifying linguistic phenomena causing the errors, or focusing on particular error type).
While the previously reported findings on this method are promising, no systematic comparison for the purposes of evaluating machine translation systems has been carried out so far. Furthermore, the simplified error marking method does not solve the challenges in determining how to appropriately weight individual errors to obtain segment-level scores, a problem that becomes particularly pronounced when extending the evaluation to document level.

This work combines the advantages of error annotation (like MQM) and assigning direct scores (like DA).
The annotators are first asked to identify and mark all errors, and afterwards to assign an overall score.
When deciding about the score, they are \textit{primed} by the preceding error annotation and see all the marked errors that can be taken into consideration for the final score.

\section{Comparison with DA+SQM and MQM}\label{sec:da+mqm}

Our proposed method lies between DA+SQM and MQM protocols, so we provide a detailed comparison between the two before describing ESA in details in the next section.

While both DA+SQM and MQM generally exhibit low inter-annotator agreement \citep{Knowles_Lo_2024, freitag-etal-2021-experts}, DA+SQM scores have high variance, which needs to be compensated with higher number of annotations per system \citep{wei-etal-2022-searching}.
On the other hand, MQM requires human experts trained with the MQM protocol and error classification.
Trained experts can be twice as expensive as translators or bilingual speakers evaluating DA+SQM. 
The required expertise is a hard constraint which makes evaluation on some languages prohibitively expensive or not possible at all, especially low-resource languages.
Furthermore, assigning a DA+SQM numerical score to a segment is anecdotally and intuitively much faster than MQM, where the evaluators need to mark each error span, classify it and assign severity.
This altogether can make each MQM annotated segment up to approximately 10$\times$ more expensive.

DA+SQM is usually based on sentence-level scores, and the paragraph-level score is computed as the average of all sentence scores in the paragraph.
Paragraph-level DA+SQM evaluation is possible, but evaluating an entire paragraph takes more time, substantially decreases the total number of collected scores, and is more demanding cognitively, which negatively impacts the inter-annotator agreement \citep{castilho-2020-page}.
On the other hand, MQM as an error classification annotation is agnostic to the choice of annotation unit.

As an example, in \ACref{fig:dasqm_mqm} we show the system ranking based on two approaches DA+SQM and MQM on Chinese-English (sentence-level evaluation) and English-German (paragraph-level evaluation) language pairs.
Although both techniques reach same order of system clusters, DA+SQM produces much fewer clusters in paragraph-level setup, thus putting many systems within a single cluster.
On the other hand, MQM is better able to distinguish different systems.
To increase statistical power of DA+SQM, we would have to collect much more DA+SQM samples \citep{wei-etal-2022-searching}, which would further drive the cost up.
In addition, DA is much more skewed towards fluency as opposed to adequacy \citep{martindale-carpuat-2018-fluency}.

The cost difference was one of the main reason behind the high usage of DA+SQM at the WMT General MT shared task \citep{kocmi-etal-2023-findings, kocmi-etal-2022-findings}.
For all these reasons, our hypothesis is that a new annotation protocol, ESA, which is between DA+SQM and MQM can provide better annotations than DA+SQM at a lower cost than MQM.

\begin{table}[tbp]
\resizebox{\linewidth}{!}{
\begin{tabular}{l>{\hspace{-5mm}}c}
\toprule
\bf Source/Translation+ESA & \bf Score \\
\midrule
\texttt{\small \bf SRC}: ... I've entered the burrata dimension. & \multirow{2}{*}{70\%} \\
\texttt{\small \bf TGT}: ... ich \hlc[red!20]{habe} die Burrata-Dimension \hlc[red!20]{eingegeben}. \\
\texttt{\small \bf gloss}: \textit{habe eingegeben(=I put in)} should be \textit{bin eingetreten}\hspace{-2cm} \\[1em]
\texttt{\small \bf SRC}: Not like other tomb raider games & \multirow{2}{*}{35\%} \\
\texttt{\small \bf TGT}: Nicht wie andere \hlc[red!50]{Gräberüberfäller} Spiele \\
\texttt{\small \bf gloss}: \textit{Gräberüberfäller} overtranslates \textit{Tomb Raider}\hspace{-2cm} \\[1em]
\texttt{\small \bf SRC}: (PERSON2) Yeah, so just know like- & \multirow{2}{*}{86\%} \\
\texttt{\small \bf TGT}: Ja, also weißt du einfach... \hlc[red!20]{[missing]} \\
\texttt{\small \bf gloss}: \textit{PERSON2} is missing \\[1em]
\texttt{\small \bf SRC}: All collards, kale, chard is transplanted. & \multirow{2}{*}{17\%} \\
\texttt{\small \bf TGT}: Alle \hlc[red!50]{Kohlköpfe}, Grünkohl, \hlc[red!50]{Schmalz} sind verpflanzt. & \\
\texttt{\small \bf gloss}: \textit{Kohlköpfe(=cabbages)} and \textit{Schmalz(=lard)} are\\ \hspace{8mm} incorrect translations \\
\bottomrule
\end{tabular}
}
\vspace{-2mm}
\captionof{example}{ESA-annotated examples with associated manual score. The error severity distinction is between \hlc[red!20]{minor} and \hlc[red!50]{major}. The first example has a single error (confusion \textit{eingeben(=put in)} as \textit{enter} meaning \textit{go in}) which also affects the auxiliary verb \text{habe/bin}. Thus, the same error is marked twice.
}
\label{tab:esa_examples}
\vspace{-2mm}
\end{table}

\section{Error Span Annotation}
\label{sec:esa}

\paragraph{Annotation process.}
In Error Span Annotation (ESA), the evaluators first mark all problematic parts (characters, words, phrases, sentences) in the translated text.
For each marked span, they also provide one of the two severity levels: \textbf{major} (e.g.~changed meaning) or \textbf{minor} (e.g.~incorrect grammar, style; see \Cref{tab:esa_examples}). 
Because all error spans are marked in the translation, not the source text, we include a special tag for marking omission errors.
This was an intentional choice over annotating the source text to make the annotation protocol forward-compatible with other translation modalities, such as audio and video translations.

\begin{figure*}[t]
    \centering
    \includegraphics[width=1\linewidth]{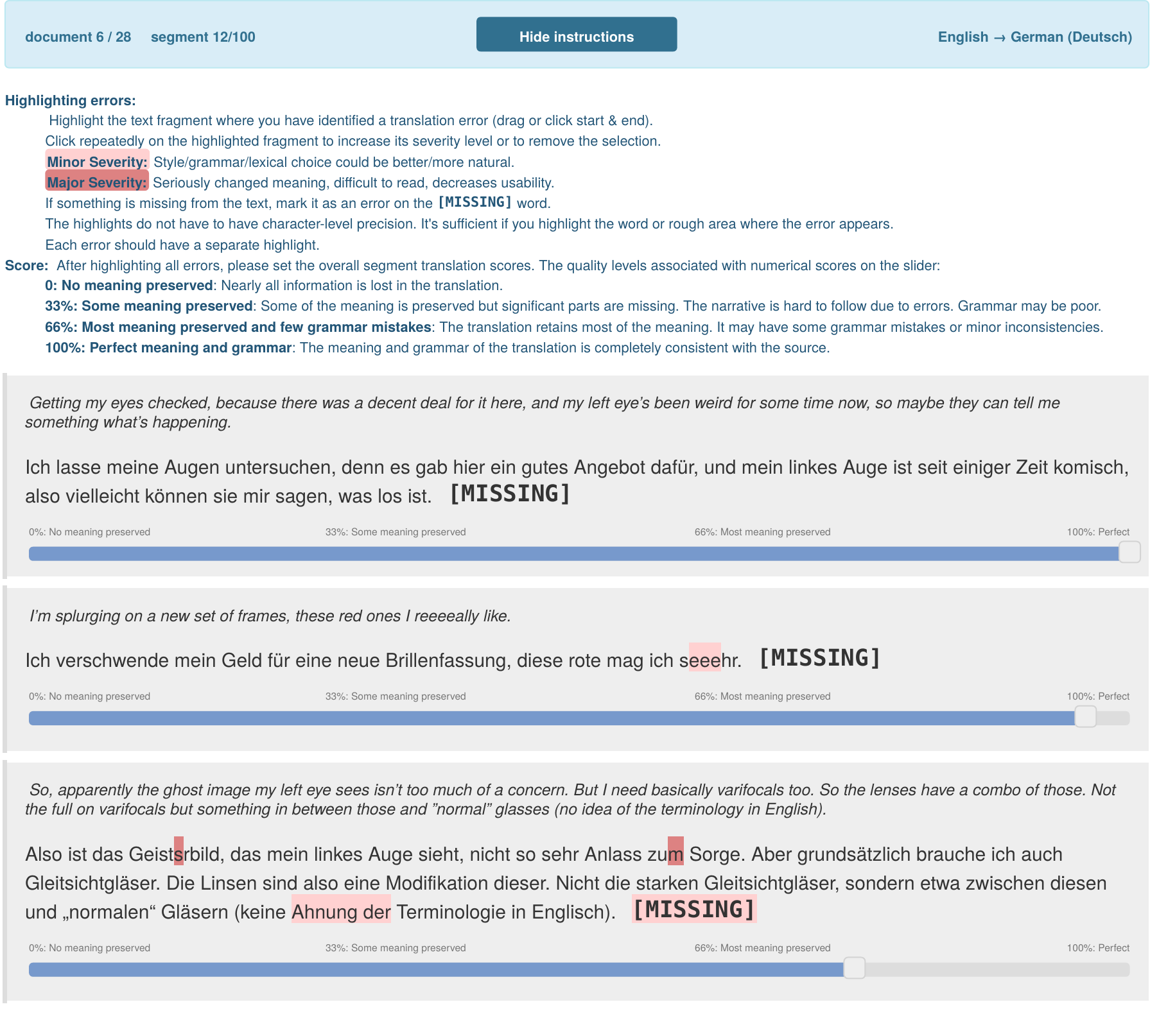}

    \vspace{-3mm}
    \caption{
    Screenshot of the beginning of one annotated document in the ESA interface (following segments are not shown).
    By showing and annotating whole documents at the segment-level, the annotators see all the relevant context.
    Segment \textit{reset} button and \textit{completed} labels removed for brevity.
    See the interactive tutorial shown to all annotators in Appendix~\Cref{fig:appraise_tutorial}.
    }
    \label{fig:appraise_screenshot_1}
\end{figure*}

After the annotators mark all the error spans, they are asked to provide an overall score for the entire segment, on the scale from 0 to 100, reminiscent of DA+SQM.
We implement the annotation interface in Appraise \citep{federmann-2018-appraise} and show a screenshot in \Cref{fig:appraise_screenshot_1}.
The full guidelines displayed to annotators are shown in \Cref{sec:user_guidelines}.

\paragraph{Segment-level scores.}

To rank systems, we need scalar values.
There are two evident ways to extract them from ESA: (1) using the annotator's overall segment-level score directly, like DA+SQM, or (2) converting error span severity levels into a segment-level score, like MQM.
By instructing the annotators to identify and mark all errors first, we \textit{prime} them to be more accurate when assessing the overall quality of the segment---when making the decision about the score, they have already marked all errors in the segment and can see them, therefore they can take them into consideration.

MQM is primarily error diagnostics protocol which has been repurposed for translation segment scoring.
The transition from MQM error spans into a single segment score has been proposed to be done with the formula based on error severity counts \citep{freitag-etal-2021-experts}:
\vspace{-1mm}
\begin{gather}
\text{MQM-like} = -5\cdot \textsc{\#major}-1\cdot \textsc{\#minor} \nonumber
\end{gather}

\noindent
Notice that this does not scale with different text sizes.\footnote{Another weighting was proposed by \citet{burchardt-2013-multidimensional} that is less prevalent in machine translation evaluation.}
As an example, if translation of a segment has two major errors, it receives the score of $-10$.
However, if the source is repeated twice and the corresponding translation as well, the score would be further decreased to $-20$.
This is especially problematic for paragraph-level evaluation which features segments of different length. 
Additionally, the segment-level MQM score might not correspond with the segment-level translation quality, such as when marking one error affecting several places in the segment, as in \Cref{tab:esa_examples} (top).
To avoid such issues, we use the annotators' direct scores as the main scoring approach for the ESA protocol unless specified differently.
The system-level scores for all human evaluation methods in this work are calculated as the average of all segment-level scores for particular system.
We further revisit the score computation in \ACref{sec:predicting_esa_score}.
In some analysis, we use ESA error spans to calculate MQM-like score, we refer to such scores as \esaspans.

\paragraph{Advantages.}
Assigning overall scores is guided by errors in the translation, and through error marking, the annotator can first focus on direct highlighting of these issues instead of determining the overall score directly.
The advantage of ESA over error classification is that it is less demanding, while still informative---the annotations can be further refined in subsequent analyses. %
Furthermore, the evaluators are not limited to any pre-defined annotation protocol and can highlight a larger range of errors. 
The error marking approach can be seen as \textit{descriptive} (encouraging annotator subjectivity and capturing their 
individual beliefs) and the error classification as a \textit{prescriptive} (discouraging annotator subjectivity and asking annotators to align with one specific belief, in this case the pre-defined error protocol), as per \citet{rottger-etal-2022-two}.

\section{Experimental setup}

We conduct experiments comparing the ESA protocol with MQM and DA+SQM protocols.
We design them in a way which makes it comparable with the previously collected annotations for \mqmwmt and \dasqmwmt. 
For this reason, we reproduce the human evaluation campaign for the WMT23 English to German systems \citep{kocmi-etal-2023-findings, freitag-etal-2023-results} with ESA and our reimplementation of MQM.

The original campaign featured 13 translations of 557 source segments.
To facilitate running multiple campaigns for proper comparison, we had to scale down and subsampled 207 segments per system (74 documents), which yields 2,691 segments in total.
To keep the ESA annotation comparable to other protocols, we subsample by selecting a subset of documents evaluated by \citet{freitag-etal-2023-results} keeping the entire documents.
This differs from \citet{freitag-etal-2023-results}, who removed some paragraphs from the ends of long documents.
In our analysis we only consider segments overlapping with both previous annotation collections, thus obtaining 2,027 annotations evaluated across all protocols.
We note that the subsampling makes all annotations protocols statistically less powerful, but keeps them fair in terms of statistical power per evaluated segment.
Therefore, clustering and final system ranking in our analysis differs from \citet{kocmi-etal-2023-findings, freitag-etal-2023-results}.
To keep the study comparable with \citet{kocmi-etal-2023-findings}, we use the Wilcoxon rank-sum test with $p{<}0.05$ when producing system clusters. However, as all systems are evaluated on the same set of segments, we advice to use Wilcoxon signed-rank test when employing ESA as proposed by \citet{kocmi-etal-2021-ship}.

\begin{table}[tbp]
\vspace{-2mm}
\centering
\resizebox{\linewidth}{!}{
\begin{tabular}{lllll}
\toprule
{} &      ESA$_1$ &      ESA$_2$ &      MQM & MQM$^\mathrm{WMT}$ \\
\midrule
\# error spans   &         0.45 &         1.00 &     0.53 &               3.37 \\
\% minor         &         63\% &         68\% &     67\% &               67\% \\
\% major         &         37\% &         32\% &     33\% &               33\% \\
Score (MQM-like) &  81.8 (-1.1) &  84.5 (-2.2) &   (-1.2) &             (-7.1) \\
\bottomrule
\end{tabular}

}
\caption{Average number of error spans per segment, ratio between minor and major errors, and scores across different annotation protocols.}
\label{tab:overview_segment_count}
\end{table}

\begin{figure}[htbp]
\includegraphics[width=\linewidth]{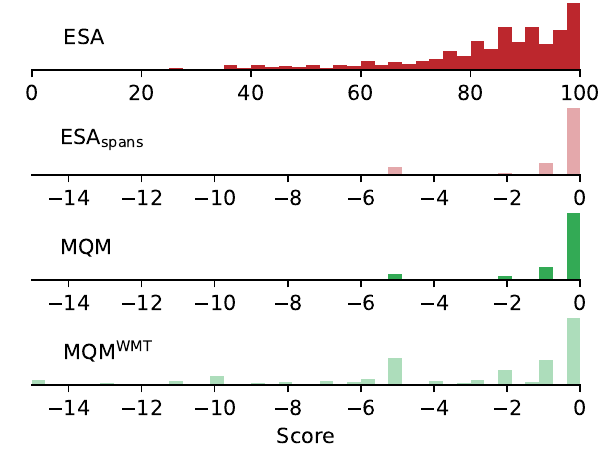}
\caption{Distribution of scores for one annotation campaign. For ESA, we either use the manual score or \esaspans computation based on error severities. For MQMs, the distribution is clipped $\geq-15$ for higher resolution.}
\label{fig:score_distribution}
\end{figure}

To analyze inter annotator agreement, we run ESA protocol twice with different sets of annotators.
We hired 28 annotators to evaluate our protocols and each protocol was evaluated by different sets of experts to avoid bias.
Specifically, we had 8 bilingual annotators for the initial run of ESA$_1$, 10 translators for ESA$_2$ (different vendor), and 10 annotators to evaluate MQM protocol.
For MQM, we hired professionals already experienced with MQM annotation protocol, while for ESA, we hired translators or bilingual speakers. 
All of them were native speakers of the target language, German.

\section{Analysis}

We first analyze the collected data, system ranking, agreement with other protocols, quality assurance, and finally the annotation time.
The findings reveal that the ESA quality is comparable, if not better than MQM, takes less time, and does not require highly trained annotators.

\subsection{Score distribution}
\label{sec:score_distribution}

As per \Cref{tab:overview_segment_count}, on average the ESA$_1$ annotators mark 0.45 error spans per segments, which is close to MQM's 0.53 error spans per segment.
The second run of ESA$_2$ has more than double of error spans per segment, which could be the result of different characteristics of annotators group, specifically experienced annotators in ESA$_1$ versus translators in ESA$_2$. 
On the other hand, \mqmwmt contains 7x more errors per segment than our re-run of MQM.
We attribute this difference primarily to the differences in annotation crowds, which further motivates our own evaluation of both MQM and ESA so that the annotations differ only in the annotation protocol and past MQM training.
The severity levels are distributed similarly across campaigns.
Important insights are in the score range distribution presented in \Cref{fig:score_distribution}: the MQM-like score computation creates more skewed distribution around 0, which is in addition unbounded and can go to -infinity the longer the evaluated segment is, complicating modeling and comparisons.
In contrast, the manual scores from annotators are spread out and guaranteed to be in [0, 100].

\begin{figure*}[t]
\includegraphics[width=\textwidth]{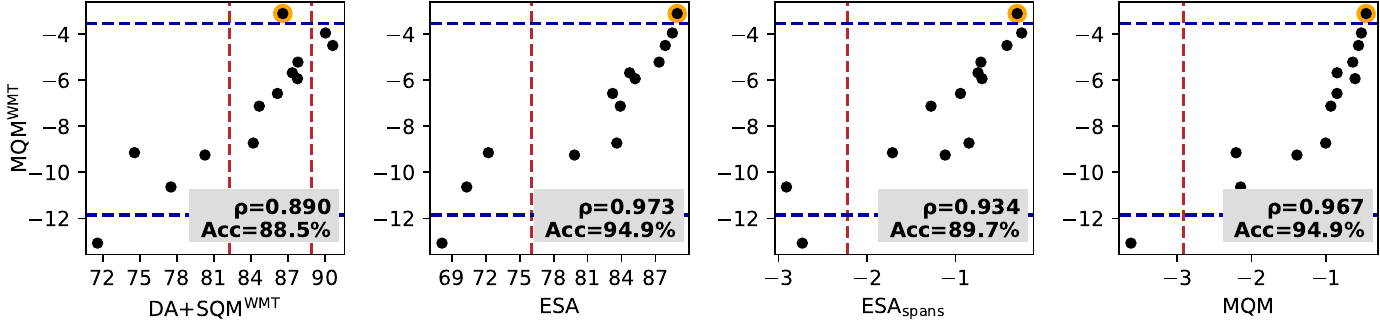}
\vspace{-7mm}
\caption{Each point represents a system, with the original \mqmwmt scores on the \textit{y}-axis plotted against our rerun of \dasqmwmt (first plot), ESA (second plot), \esaspans (third plot), and MQM (forth plot). Stripped lines indicate cluster separations determined by each method with alpha threshold 0.05. We compute Spearman correlation $\rho$ and pairwise accuracy $\textsc{Acc}$.}
\label{fig:clusters_esa}
\vspace{-3mm}
\end{figure*}

\subsection{System ranking capabilities}

We now investigate how well different annotation protocols (ESA and MQM) can rank MT systems.
For purpose of this experiment, we consider \mqmwmt which comes from an independent very high-quality annotation crowd and implementation, as the gold standard.
Note that this creates a positive bias towards our implementation of MQM as opposed to ESA.
In \Cref{sec:subset_consistency} we show an evaluation without this gold standard assumption.

When comparing two protocols, we ideally want them to rank all systems in the same order. This is not always possible as some systems are very similar and cannot be significantly distinguished with the evaluated sample size. Another problem is that different protocols may weight different phenomena (e.g.~fluency or adequacy) differently.
To compare different protocols in the task of ranking systems, we use pairwise accuracy \citep{kocmi-etal-2021-ship}, which is also used in WMT Metrics shared task when comparing different automatic metrics \citep{freitag-etal-2023-results}.
Pairwise accuracy measures how many system pairs does a protocol rank the same way as \mqmwmt.
As we have only 78 system pairs, any wrong system pair will change pairwise accuracy by 1.28\%.
Therefore, we also calculate Spearman's correlations as we mainly want protocols to have monotonic ranking.

In \Cref{fig:clusters_esa}, each subplot compares system-level scores between one protocol on x-axis and \mqmwmt on y-axis.
Our repeated MQM experiment and ESA protocol rank systems identically (94.9\%), while ESA has slighly higher Spearman's correlation with \mqmwmt.
On the other hand, \dasqmwmt significantly lacks behind both protocols. 
This suggest that our ESA protocol has comparable system ranking capabilities to MQM and is superior to \dasqmwmt.

The \Cref{fig:clusters_esa} also shows, that relying on error spans only is not optimal, as \esaspans{} has lower accuracy and Spearman's correlation than ESA. We can notice that this is even lower than our rerun of MQM. This can be attributed to the evaluation crowd, where we used professional MQM annotators for MQM protocol, while we used bilingual speakers and translators for the ESA protocol.

Further focusing on the clustering, \mqmwmt significantly differentiates the top system from others (highlighted in orange), while \dasqmwmt strongly puts this system into the second cluster.
This system is human reference, which we assume should be of highest quality.
The reduced number of clusters in contrast to \Cref{fig:dasqm_mqm} is due to the lower sample size.
This may be result of \dasqmwmt higher sensitivity to fluency and style errors, which contribute to 60\% of all errors in human reference as marked by \mqmwmt.
This conflict in clustering is one of the critiques of \dasqmwmt if we assume that human reference should be the highest scoring translation.
For example, in \citet{kocmi-etal-2023-findings}, human reference was the best translation only in 2 out of 8 language pairs.

\begin{table}[htbp]
\centering
\resizebox{0.7\linewidth}{!}{
\begin{tabular}{ll}
\toprule
{} & MQM$^\mathrm{WMT}$ \\
\midrule
ESA$_1$               &              0.227 \\
ESA$_1$$_\mathrm{\,spans}$       &              0.170 \\
ESA$_2$               &              0.250 \\
ESA$_2$$_\mathrm{\,spans}$       &              0.236 \\
MQM                   &              0.189 \\
DA+SQM$^\mathrm{WMT}$ &              0.209 \\
\bottomrule
\end{tabular}

}
\caption{Kendall $\tau$ segment-level correlations between evaluation protocols.}
\label{tab:cross_protocol_correlation}
\end{table}

\subsection{Agreement with other protocols}

We now compare how different protocols correspond on the segment-level.
We use \mqmwmt as the gold standard to compare against because it was done independently outside of our setup and with high quality assurance.
We analyze two aspects: (1) segment scores, and (2) spans, where we consider spans overlapping even with a single character as a match.
For this evaluation, we use \href{https://en.wikipedia.org/wiki/Kendall_rank_correlation_coefficient#Tau-c}{Kendall $\tau$ variant C}, which is more suitable for data with different underlying scales and many ties.

In \Cref{tab:cross_protocol_correlation}, we see that although all protocols correlate similarly with \mqmwmt, our protocols obtain the highest $\tau$ for both runs. The segment-level correlation also confirms that relying only on the error spans is not optimal and \esaspans obtains lower Kendall score.

In \Cref{tab:quality_of_esa_spans} we focus on the annotated spans.
We consider any spans that overlap as matching, irrespective of severity.
Because different protocols have different average number of error spans, presenting just the size of the intersection would be misleading.
Instead, we show normalized set similarity that is not symmetric.
It answers the questions: \textit{What proportion of samples in $B$ were covered by $A$?}
Both ESA and MQM cover \mqmwmt similarly, with 29\% and 32\% respectively.
At the same time, \mqmwmt covers ESA and MQM with 93\% and 85\%. 

\begin{table}[ht]
\centering
\resizebox{0.85\linewidth}{!}{
\begin{tabular}{lccc}
\toprule
$\bm{\downarrow}$A \hfill B$\bm{\rightarrow}$ \hspace{1mm}
& ESA & MQM & \mqmwmt \\
\midrule
ESA &  & 77\% & 29\% \\
MQM & 85\% &  & 32\% \\
\mqmwmt & 93\% & 89\% & \\
\bottomrule
\end{tabular}
}
\begin{center}
$\bm{|A\cap B| / |B|}$ \quad {\small (100\% for $B=\emptyset$)}
\end{center}
\vspace{-2mm}
\caption{
Similarity between spans of different annotation protocols (one campaign) computed as percentage how much of $B$ does $A$ contain.
For example, 93\% of ESA spans were also in MQM.
Any span that overlaps with another one is considered a hit, even if only with as single character.
}
\label{tab:quality_of_esa_spans}
\end{table}

\subsection{Quality of annotations}

\paragraph{Intra annotator agreement.}

We want the human evaluation protocol to consistently assign similar scores for the same translations over time.
A good indicator of the annotation quality is how noisy and subjective it is, which is reflected by how much annotators agree on the same segments (inter annotator agreement) as well as how a single annotator agrees with themselves (intra annotator agreement).
To measure intra-AA, we ask the same annotators to again annotate the same documents two months later.
We prepare an identical campaign with the same distribution of systems in the same order as originally, asking the same annotators to redo it again for both ESA and MQM.

It is not obvious how to measure the agreements for protocols that have different features.
One issue is the frequency of ties, where MQM has more ties than rating from ESA or \dasqmwmt.For example, \mqmwmt contains 30.8\% no-errors, while \dasqmwmt contains only 5.2\% of score 100.
Secondly, each protocol uses different range and distribution of scores, which makes calculation of agreement complicated. See \ACref{fig:intra_annotator_agreement} to understand different distribution of scores.

Previous works comparing different protocols discretize the scale into bins \citep{graham-etal-2013-continuous, freitag-etal-2021-experts}, however, this approach is sensitive to subjective selection of bin sizes and benefits already discrete protocols.
Instead, we propose to use Kendall's Tau-c correlation to measure inter-annotator agreement.
Secondly, we also want to take into consideration that small changes in scores are less damaging than large shifts, therefore, we want inter annotator's scores to correlate linearly, ideally having identical score each time.
To measure this, we use Pearson's correlation.
Lastly, we measure recall of how often annotator mark \textit{any} error in the same segment in contrast to leaving the segment without marked errors.

\Cref{tab:intra_annotator_agreement} shows that ESA has all scores higher than MQM. 
Higher Kendall and Pearson suggest that the task is easier for annotators to agree on the score.
We expect the recall to be comparable for both techniques as the task is similar, we hypothesize that the drop in MQM could be explained by annotators saving time and skipping minor errors as the annotation is more complicated for MQM than ESA, which can be confirmed when looking at minor error's recall only.

Lastly, evaluating inter annotator agreement is heavily affected by the strategy of annotators, where different annotation strategy does not mean different performance of the task as \citet{riley2024finding} showed. Secondly, the \mqmwmt was collected with a different tooling and the documents have been presented to annotators in different order, which could also impact the inter annotator agreement.

\begin{table}[htp]
\centering
\resizebox{0.95\linewidth}{!}{
\begin{tabular}{l|ll|ll}
\toprule
{} & \multicolumn{2}{l}{Intra AA} & \multicolumn{2}{l}{Inter AA} \\
{} &      ESA &     MQM &      ESA &     MQM \\
\midrule
Kendall's Tau-c &    0.149 &   0.109 &    0.254 &   0.116 \\
Pearson         &    0.403 &   0.189 &    0.482 &   0.281 \\
Error recall    &   69.6\% &  61.9\% &   66.6\% &  40.1\% \\
Minor e. recall &   70.7\% &  66.2\% &   67.7\% &  44.4\% \\
Major e. recall &   82.6\% &  82.1\% &   84.8\% &  62.9\% \\
\bottomrule
\end{tabular}

}
\vspace{-1mm}
\caption{Intra- and inter-annotator agreement on segment-level.}
\label{tab:intra_annotator_agreement}
\vspace{-2mm}
\end{table}

\paragraph{Inter annotator agreement.}

To measure how different annotators agree between themselves on the same protocol, we compare ESA$_1$ to ESA$_2$, where each protocol was evaluated by different group of annotators (bilingual annotators vs. translators).
To calculate MQM's inter-annotator agreement, we compare our MQM run with \mqmwmt.
However, the comparison is not as 1:1 as for ESA protocol.
Our MQM was collected with different interface and system outputs have been shown to annotators in a different order.
Results in \Cref{tab:intra_annotator_agreement} suggest, that ESA has also higher inter-annotator agreement than MQM.
Unfortunately, we could not rerun DA+SQM protocol to calculate agreements, which needs to be reevaluated in future work.

\paragraph{Agreement on the error span.}
We now investigate how much MQM annotators agree on the error spans, error categories and severity levels. We evaluate from two angles: \textit{intra}, where we check if the same annotator marks the same error spans or have overlapping parts, and \textit{inter}, where we compare our MQM error spans to \mqmwmt.
\Cref{tab:mqm_categories} shows that only in 30\% of cases, the same annotator marked at least part of the same segment as an error regardless error severity and category. If we look at cases preserving severity and category, this number drops to 8.4\%.

When comparing the inter annotator agreement, only 50\% of errors are overlapping. This number is higher as the total number of errors in \mqmwmt is 7$\times$ higher than in MQM, therefore it is more likely an error will have overlap.

\begin{table}[htbp]
\centering
\resizebox{0.95\linewidth}{!}{
\begin{tabular}{lll}
\toprule
{} &  Intra AA & Inter AA \\
\midrule
Any errors            &  29.3\% &            50.2\% \\
Same severity         &  16.9\% &            23.7\% \\
Same category         &  18.9\% &            24.1\% \\
Same sev. + categ.    &  11.6\% &            10.0\% \\
Same sev. + subcateg. &   8.4\% &                 - \\
\bottomrule
\end{tabular}

}
\caption{ESA intra- and inter-annotator agreement (frequency) on marking overlapping errors with same severities, categories or subcategories.}
\label{tab:mqm_categories}
\end{table}

\paragraph{Quality control.}
To measure the quality of annotations, we added ``attention checks'' in the form of segments for which we can reliably check whether the annotator annotated them correctly or not.
In random documents, we perturbed the translation by replacing part of it with random sequence of words of the same length  introducing a major translation error.
Within 100 segments annotated by the annotator, they see both original and perturbed versions of that document.
We can control the annotation quality by checking if the perturbed documents received more error spans.
We show a worked-out \Cref{ex:attention_chec}.

\begin{figure}[htbp]
\centering
\begin{minipage}{0.93\linewidth}
\small\it
\texttt{\bf SRC}:\hspace{2mm} Sie haben gestern das Treffen wieder verschoben. \\
\texttt{\bf TGT}:\hspace{1.5mm} He postponed the meeting again yesterday. \\
\texttt{\bf TGT$^\textbf{P}$}: He postponed the meeting \ul{squirrels tense}.
\end{minipage}
\captionof{example}{An example of a perturbed translation \texttt{\bf TGT$^\textbf{P}$} based on the original system translation \texttt{\bf TGT}. The \texttt{\bf TGT} has one error (\textit{He} should be \textit{You} or \textit{They}) and \texttt{\bf TGT$^\textbf{P}$} introduces one more errors (\textit{squirrels tense}).}
\label{ex:attention_chec}
\end{figure}

The MQM and ESA setups used the same perturbations and we show their results in \Cref{tab:attention_checks}.
The scores comparing original and perturbed segments for ESA and MQM are vastly different,\footnote{The protocols use different scales, for example, one major error under MQM is $-5$ points, while 25 points in ESA represents quarter of the full scale.} showing that annotators paid attention to the quality control items.
For ESA the original segment had a higher score than the perturbed one in 86\% of cases, while for MQM in 78\% of cases. When investigating whether an annotator marked the error span or not, MQM has higher recall than ESA, however, this is less crucial for ESA as annotator can adjust the ranking without marking the error.

\begin{table}[ht]
\resizebox{\linewidth}{!}{
\begin{tabular}{llccc}
\toprule
& & \bf Original & \bf Perturbed & \bf OK \\
\midrule
\multirow{3}{*}{\bf ESA} & Score & 79.5 & 52.6 & 86\% \\
& Span count & 0.85 & 1.86 & 54\% \\
& Perturbation marked\hspace{-20mm} & & & 56\% \\[0.7em]
\multirow{3}{*}{\shortstack[l]{\bf MQM}} & Score & -1.87 & -6.49 & 78\% \\
& Span count & 0.66 & 1.68 & 70\% \\
& Perturbation marked\hspace{-20mm} & & & 76\% \\
\bottomrule
\end{tabular}
}
\caption{Annotations assigned to perturbed attention check items (either scores or number of spans).
\textbf{OK} is percentage in how many cases the non-perturbed item received a higher score or had fewer error spans, and how often the perturbed span was marked by the annotator.}
\label{tab:attention_checks}
\vspace{-2mm}
\end{table}

\subsection{Annotation time}

One of the reasons behind development of the new protocol is reducing the time requirements of the evaluation.
In this section, we analyze times on our experiments only, therefore our rerun of MQM in contrast to ESA.
We do not include the times for DA+SQM and \mqmwmt because they were done independently outside of our study and the time data are not available.
Assessing the speed of annotations is challenging as annotators took breaks during the annotation (from short breaks taking several minutes up to several hours).
This makes the evaluation of time problematic and we therefore investigate the time estimate in several ways.
Counting all annotators together, the median time for annotation per single paragraph for MQM is 38 seconds and for ESA is 29 seconds, a reduction of 23\%.
However, as median time for each annotator fluctuates, we look at the average median time across annotators.
\textbf{For MQM, the median is 49 seconds and for ESA it is 34 seconds, a reduction of 32\%}.
This can be contributed mainly to the less demanding error span annotation approach used in ESA.

We note that further speedups could be made by instructing the ESA annotators to not spend extra time e.g. marking multiple error-span annotations of a single grammatical phenomenon.

\paragraph{Speedup during annotations.}

\begin{figure}[htbp]
\includegraphics[width=\linewidth]{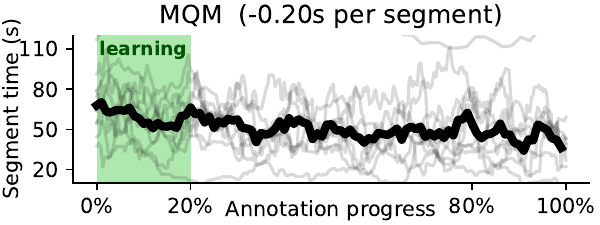}
\vspace{-3mm}

\includegraphics[width=\linewidth]{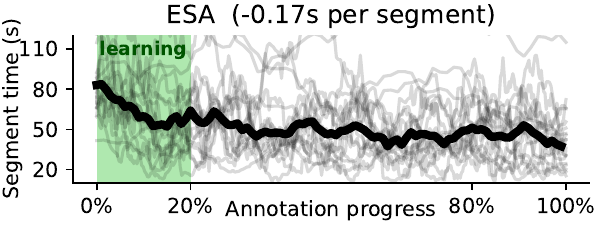}
\caption{Time per segment with respect to progression in the annotation. The faint gray lines represent individual annotators, while the bold black line shows the average time. The lines are smoothed with a window of size 15 segments. We also compute the average speed at the beginning and at the end, which yields the \textit{learned speedup}. This is how much the annotator speeds up after working on one segment.}
\label{fig:document_speedup}
\vspace{-2mm}
\end{figure}

Naturally, the reported total annotation time does not distinguish between the duration of the first and last annotations.
In practice, annotators \textit{learn} to perform the annotation task more effectively.
In \Cref{fig:document_speedup} we show time per segment depending on how many segments the annotator already processed.
For both MQM and ESA, there is a small learning effect.
For MQM, with each segment, the annotator becomes 0.20s faster, while for ESA this is 0.17s.

\section{Conclusion}

Existing annotation protocols for machine translation evaluation are either expensive because they require expert labor (MQM), or they are noisy and less reliable (DA+SQM).
To this end, we propose, describe, and analyze \textbf{Error Span Annotation (ESA)}, which builds on top of previous protocols to enable economic evaluation at scale.
It works by asking the annotators to mark error spans, but with only the error severities and not types.
In contrast to MQM, we also solicit final translation score.
This is more reliable than DA+SQM, as the annotators are primed and informed about the translation errors to assess quality of longer documents.

We showed that our protocol has the higher inter and intra annotator agreement than MQM while being 32\% faster. In addition, the protocol does not require annotators trained in MQM categorizations.

Lastly, we showed that relying only on error spans and not using the ranking score as we did in \esaspans produces suboptimal scoring, therefore the combination of error spans and ranking seems to produce the best results.

\section*{Limitations}

A possible limitation in contrast to MQM is that now the system evaluation does not provide breakdown of error types, which could help practitioners in improving their systems.
ESA does not provide this because the goal is \textit{evaluation} and not \textit{diagnosis}.
Because of this and the costs of scaling expert labor, we are convinced that this is not a true shortcoming of ESA.
Furthermore, the annotated errors can be further classified and analysed, if necessary. 

Our experiments, for monetary reasons, were done only on one language pair, English$\rightarrow$German.
Nevertheless, it is unlikely that the results would be vastly different for other languages.
The most difficult setup could be with Chinese, Japanese, and Korean texts that do not use spaces.
However, we made a deliberate decision to allow highlighting of individual characters, as opposed to words, so that the user experience is unified across all languages.
This was done in spite of speed improvements (selecting on the word level is easier than selecting individual character boundaries) in order to make the tool scalable to a large range of languages.

\section*{Ethics Statement}

The annotators were paid a standard commercial translator wage in the respective country. The experts in the MQM annotation has been paid double the hourly wage.
No personal data was collected and the showed data was screened for potentially disturbing content.

We follow up with a questionnaire asking annotators on their feedback. Almost all annotators specified that the annotation experience was positive and instructions were clear. The main concern they mentioned was that some documents have been too long to evaluate.

\vspace{-3mm}

\bibliography{misc/anthology.min.bib,misc/bibliography.bib}
\bibliographystyle{misc/acl_natbib}

\clearpage

\appendix
\section{User Guidelines}
\label{sec:user_guidelines}

The following are annotation guidelines for our local ESA and MQM campaigns.

\subsection{\hspace{-1mm}ESA (Error Span Annotations)}

{
\fontsize{10}{11}\selectfont

\paragraph{Higlighting errors:}
Highlight the text fragment where you have identified a translation error (drag or click start \& end).
Click repeatedly on the highlighted fragment to increase its severity level or to remove the selection.
\begin{itemize}[topsep=0mm]
\item        \textbf{Minor Severity:} Style/grammar/lexical choice could be better/more natural.
\item        \textbf{Major Severity:} Seriously changed meaning, difficult to read, decreases usability.
\end{itemize}
If something is missing from the text, mark it as an error on the \texttt{\bf [MISSING]} word.
The highlights do not have to have character-level precision. It's sufficient if you highlight the word or rough area where the error appears.
Each error should have a separate highlight.
        
\paragraph{Score:} After highlighting all errors, please set the overall segment translation scores. The quality levels associated with numerical scores on the slider:
\begin{itemize}[topsep=0mm]
\item 0\%: No meaning preserved: Nearly all information is lost in the translation.
\item 33\%: Some meaning preserved: Some of the meaning is preserved but significant parts are missing. The narrative is hard to follow due to errors. Grammar may be poor.
\item 66\%: Most meaning preserved and few grammar mistakes: The translation retains most of the meaning. It may have some grammar mistakes or minor inconsistencies.
\item 100\%: Perfect meaning and grammar: The meaning and grammar of the translation is completely consistent with the source.
\end{itemize}

\subsection{\hspace{-1mm}MQM (Multidimensional Quality Metrics) }

\paragraph{Higlighting errors:}
        Highlight the text fragment where you have identified a translation error (drag or click start \& end).
        Click repeatedly on the highlighted fragment to increase its severity level or to remove the selection.
\begin{itemize}[topsep=0mm]
\item        \textbf{Minor Severity}: Style/grammar/lexical choice could be better/more natural.
\item        \textbf{Major Severity}: Seriously changed meaning, difficult to read, decreases usability.
\end{itemize}
        If something is missing from the text, mark it as an error on the \texttt{\bf [MISSING]} word.
        The highlights do not have to have character-level precision. It's sufficient if you highlight the word or rough area where the error appears.
        Each error should have a separate highlight.
        
\paragraph{Error types:}
After highlighting an error fragment, you will be asked to select the specific error type (main category and subcategory).
If you are unsure about which errors fall under which categories, please consult the \href{https://themqm.org/the-mqm-typology/}{typology definitions}.

}

\begin{figure}[htbp]
\centering
\includegraphics[width=0.95\linewidth]{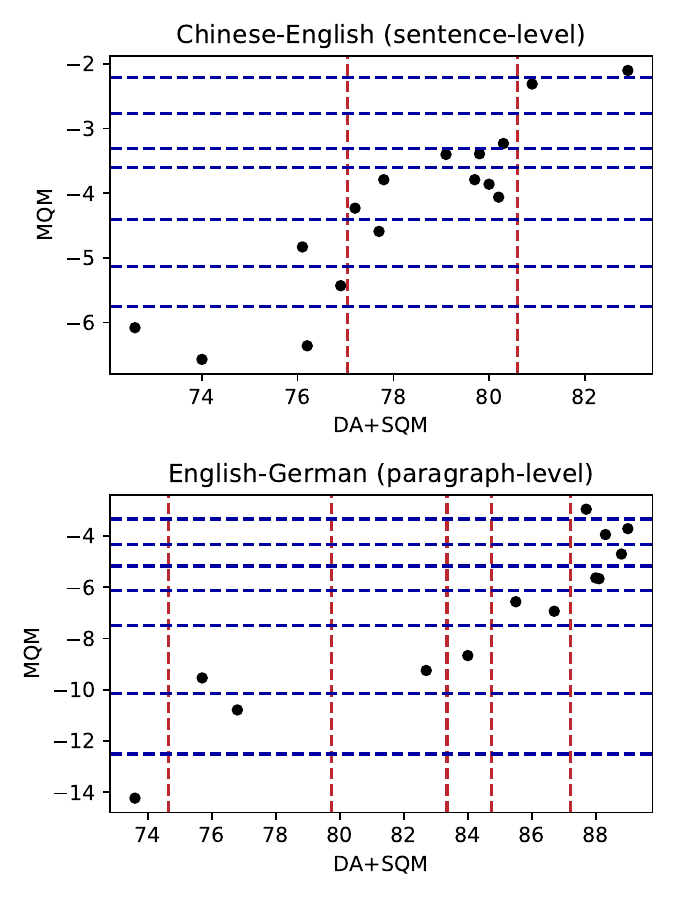}

\vspace{-3mm}
\caption{System scores illustrating differences between DA+SQM and MQM. Each point is a single system and dashed lines mark clusters. DA+SQM produces fewer clusters and groups many systems into one single cluster, while MQM better distinguishes different systems. Scores and clusters are from \citet{kocmi-etal-2023-findings}.
}
\label{fig:dasqm_mqm}
\end{figure}

\begin{figure}[htbp]
\includegraphics[width=\linewidth]{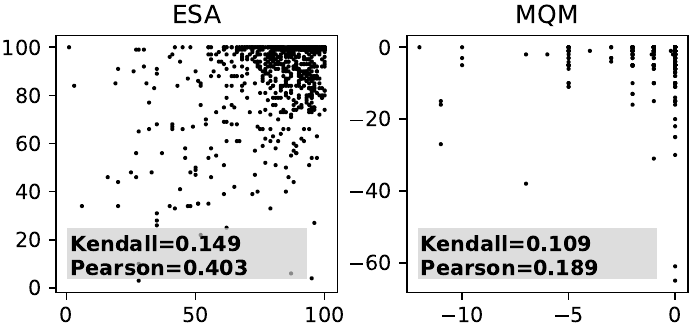}

\vspace{-3mm}
\caption{Intra annotator agreement; changes in scoring by the same annotator when evaluated again. Each point represents an annotated segment with x-axis being annotator's score assigned in March and y-axis their score assigned in May.}
\label{fig:intra_annotator_agreement}
\end{figure}

\begin{table}[htbp]
\resizebox{\linewidth}{!}{
\begin{tabular}{lc}
\toprule
\bf Feature & \hspace{-18mm} \bf Corr. with ESA score ($\bm{\rho}$) \\
\midrule
Source token count & -0.16\\
Target token count & -0.06\\
\\[-0.5em]
Minor error count & -0.20\\
Major error count & -0.52\\
Missing error count & -0.45\\
\\[-0.5em]
Minor error count (normalized) & -0.13\\
Major error count (normalized) & -0.37\\
Missing error count (normalized) & -0.31\\
\bottomrule
\end{tabular}
}

\vspace{-2mm}
\caption{Segment-level Pearson correlation of individual features with the ESA score.}
\label{tab:predicting_score_corr}
\end{table}

\clearpage

\section{Additional results}

\subsection{From error spans to final score}
\label{sec:predicting_esa_score}

To find out what influences the score, we show correlation between individual segment-level features in \Cref{tab:predicting_score_corr}.
On average, longer segments have lower translation quality.
Importantly, the error counts normalized by segment length correlate less than the non-normalized counterparts.
However, we note that the normalized scores are more continuous that the non-normalized MQM computation, as per \Cref{fig:score_distribution}.

\begin{figure}[htbp]
\centering
\includegraphics[width=0.8\linewidth]{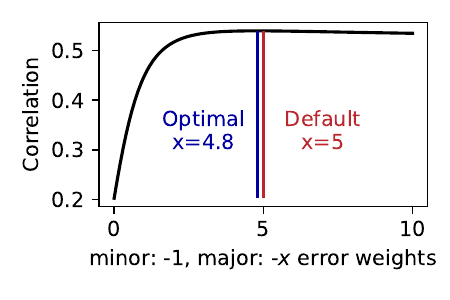}
\caption{Correlation between direct ESA score and scores computed from error spans with minor errors having weight $-1$ and major errors $-x$.}
\label{fig:predicting_score}
\end{figure}

The MQM formula was crafted with respect to preserving system ranking and not segment-level matching from the same annotator \citep{freitag-etal-2021-experts}.
However, the construction of ESA allows us to revisit this problem as each annotator gives both the error spans and the final score.
We scan for multiple minor/major ratios of error weights and show the results in \Cref{fig:predicting_score}.
We find that the optimal formula that optimizes the correlation between the direct score and the score from the spans has the following form:
$\textsc{Seg.Score} = -1\cdot \textsc{\#minor} -4.8\cdot \textsc{\#major}$,
which is very close to the originally proposed 1:5 ratio.
From \Cref{fig:predicting_score}, the weight of the major error class seems to have a much bigger effect on the final translation score, suggesting that minor errors play a lesser role.

\subsection{Protocol evaluation without\linebreak a gold standard}
\label{sec:subset_consistency}

Evaluating the quality of a protocol without a target to compare to is difficult.
In previous sections we assumed that \mqmwmt is the gold standard, which might bias the evaluation in favor of MQM.
Even though the results showed higher correlations of ESA with \mqmwmt.
For completeness, we consider the methodology of \citet{zouharkocmi2024esaai} which does not require target gold standard to compare the quality of annotation protocols.

The assumption is that annotation protocols have various levels of noise, but are unbiased in what they measure.
Because MQM and DA+SQM might measure different things, we want to compare each to the perfect ranking of the particular thing they aim to measure.
The linking hypothesis is that even noisy and low-quality annotation protocols would lead to the final system ranking with large enough data.
Vice-versa, only robust annotation protocols would arrive at the final ranking with only a few data points per each system.
This is formalized by the subset consistency accuracy.
It is the system ranking accuracy on a subset of annotations with respect to the ranking induced by the full data from one annotation protocol.

\begin{figure}[htbp]
    \centering
    \includegraphics[width=1\linewidth]{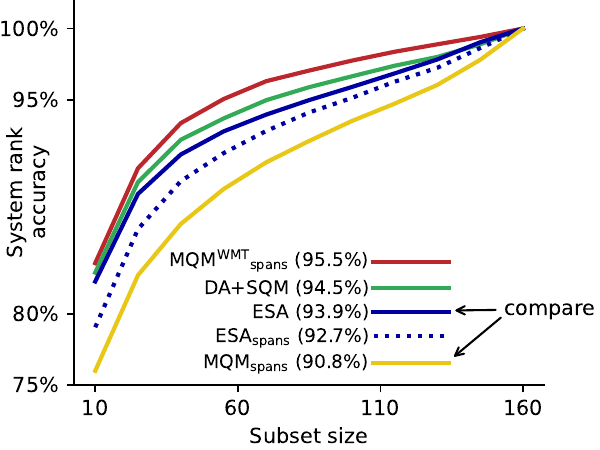}
    \caption{Subset consistency accuracy \citep{zouharkocmi2024esaai} of annotation protocols. E.g. with just 60 annotated segments, \mqmwmt achieves 95\% system ranking accuracy with its final ranking based on 160 annotated segments. Values in the legend are averages, corresponding to normalized area under the curve. The only comparable lines are ESA, ESA\textsubscript{spans}, and MQM\textsubscript{spans} because they were run in the same setting with similar crowds.}
    \label{fig:subset_consistency}
\end{figure}

We show the results in \Cref{fig:subset_consistency}.
Out of the comparable lines, ESA (with direct scoring) achieves the highest subset consistency accuracy.
In practice, this translates to needing fewer annotated examples to achieve the final system ranking.
This directly corresponds to lower annotation costs.

\begin{figure*}[htbp]
\includegraphics[width=\textwidth]{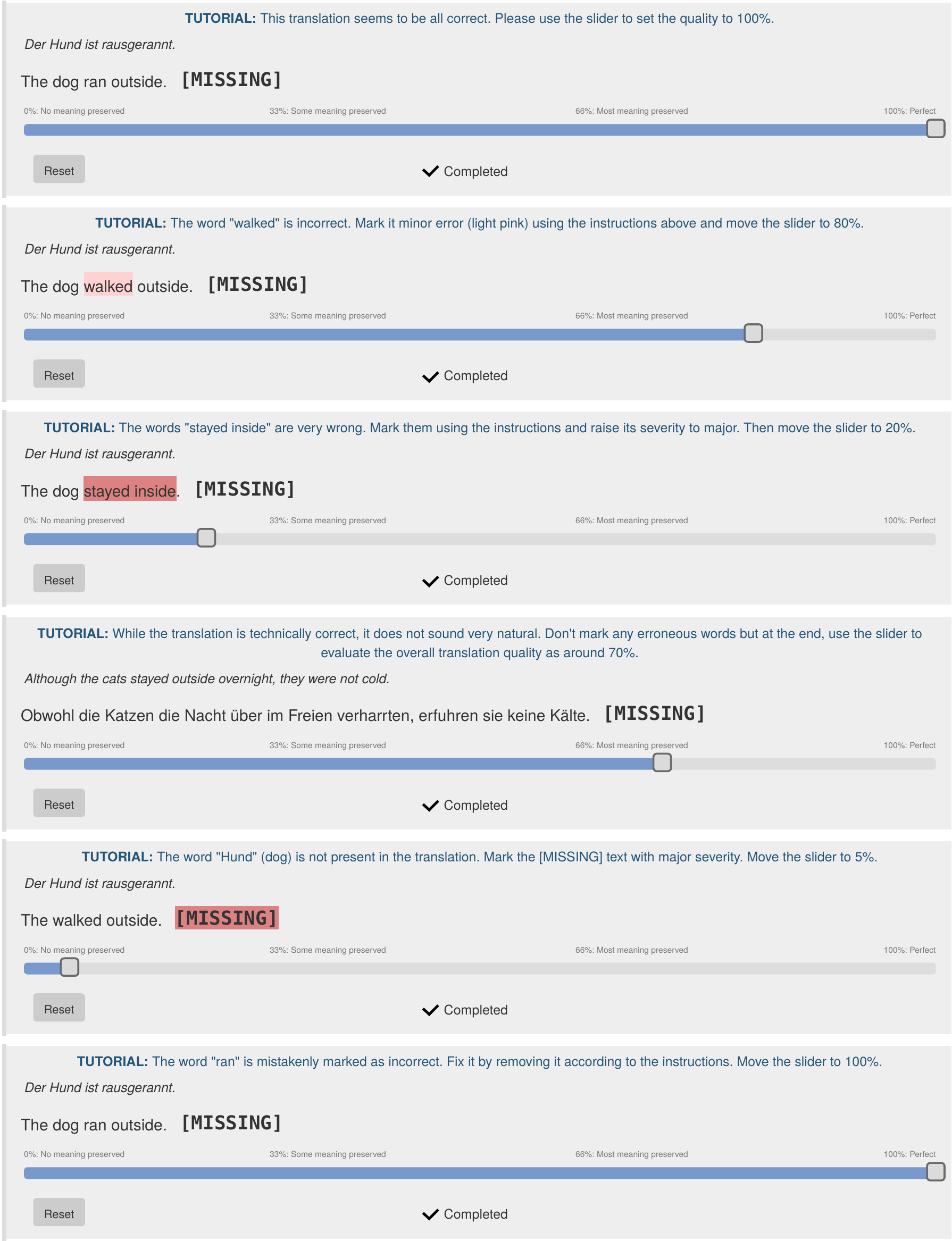}

\caption{Tutorial to ESA annotations shown at the beginning of the campaign. All tutorial segments need to be annotated correctly before continuing.}
\label{fig:appraise_tutorial}
\end{figure*}

\end{document}